\documentclass[sigplan,screen]{acmart}

\AtBeginDocument{%
  \providecommand\BibTeX{{%
    \normalfont B\kern-0.5em{\scshape i\kern-0.25em b}\kern-0.8em\TeX}}}

\copyrightyear{2021} 
\acmYear{2021} 
\setcopyright{acmcopyright}\acmConference[ICCPS '21]{ACM/IEEE 12th International Conference on Cyber-Physical Systems (with CPS-IoT Week 2021)}{May 19--21, 2021}{Nashville, TN, USA}
\acmBooktitle{ACM/IEEE 12th International Conference on Cyber-Physical Systems (with CPS-IoT Week 2021) (ICCPS '21), May 19--21, 2021, Nashville, TN, USA}
\acmPrice{15.00}
\acmDOI{10.1145/3450267.3450541}
\acmISBN{978-1-4503-8353-0/21/05}

\newcount\Comments  
\usepackage{color}
\definecolor{darkgreen}{rgb}{0,0.5,0}
\definecolor{purple}{rgb}{1,0,1}
\definecolor{teal}{rgb}{0,0.4627,0.5804}
\newcommand{\kibitz}[2]{\ifnum\Comments=1\textcolor{#1}{#2}\fi}

\usepackage[belowskip = 0pt,aboveskip=0pt]{subcaption}
\usepackage{enumitem}
\usepackage[belowskip=0pt,aboveskip=2pt]{caption}
\setlist{nosep}

\begin{document}

\title{CAN Coach: Vehicular Control through Human Cyber-Physical Systems}

\author{Matthew Nice}
\email{matthew.nice@vanderbilt.edu}
\orcid{0000-0003-3609-1760}
\affiliation{
\institution{Vanderbilt University}
 \streetaddress{1025 16th Ave S.}
 \city{Nashville}
 \state{Tennessee}
 \postcode{37212}
}
\author{Safwan Elmadani}
\affiliation{
\institution{University of Arizona}
 \city{Tuscon}
 \state{Arizona}
 \postcode{85721}
}

\author{Rahul Bhadani}
\affiliation{
\institution{University of Arizona}
 \city{Tuscon}
 \state{Arizona}
 \postcode{85721}
}
\author{Matt Bunting}
\affiliation{
\institution{University of Arizona}
 \city{Tuscon}
 \state{Arizona}
 \postcode{85721}
}
\author{Jonathan Sprinkle}
\orcid{0000-0003-4176-1212}
\affiliation{
\institution{University of Arizona}
 \city{Tuscon}
 \state{Arizona}
 \postcode{85721}
}
\author{Dan Work}
\affiliation{
\institution{Vanderbilt University}
 \streetaddress{1025 16th Ave S.}
 \city{Nashville}
 \state{Tennessee}
 \postcode{37212}
}

\renewcommand{\shortauthors}{M. Nice, S. Elmadani, R. Bhadani, M. Bunting, J. Sprinkle, D. Work}

\begin{abstract}
  This work addresses whether a \textit{human-in-the-loop cyber-physical system} (HCPS) can be effective in improving the longitudinal control of an individual vehicle in a traffic flow. We introduce the \textit{CAN Coach}, which is a system that gives feedback to the human-in-the-loop using  radar data (relative speed and position information to objects ahead) that is available on the \textit{controller area network} (CAN). Using a cohort of six human subjects driving an instrumented vehicle, we compare the ability of the human-in-the-loop driver to achieve a constant time-gap control policy using only human-based visual perception to the car ahead, and by augmenting human perception with audible feedback from CAN sensor data.  The addition of CAN-based feedback reduces the mean time-gap error by an average of 73\%, and also improves the consistency of the human by reducing the standard deviation of the time-gap error by 53\%. We remove human perception from the loop using a \textit{ghost mode} in which the human-in-the-loop is coached to track a virtual vehicle on the road, rather than a physical one. The loss of visual perception of the vehicle ahead degrades the performance for most drivers, but by varying amounts. We show that human subjects can match the velocity of the lead vehicle ahead with and without CAN-based feedback, but velocity matching does not offer regulation of vehicle spacing. The viability of dynamic time-gap control is also demonstrated. We conclude that (1) it is possible to coach drivers to improve performance on driving tasks using CAN data, and (2) it is a true HCPS, since removing human perception from the control loop reduces performance at the given control objective.
\end{abstract}



\begin{teaserfigure}
\centering
  \includegraphics[width=0.75\textwidth]{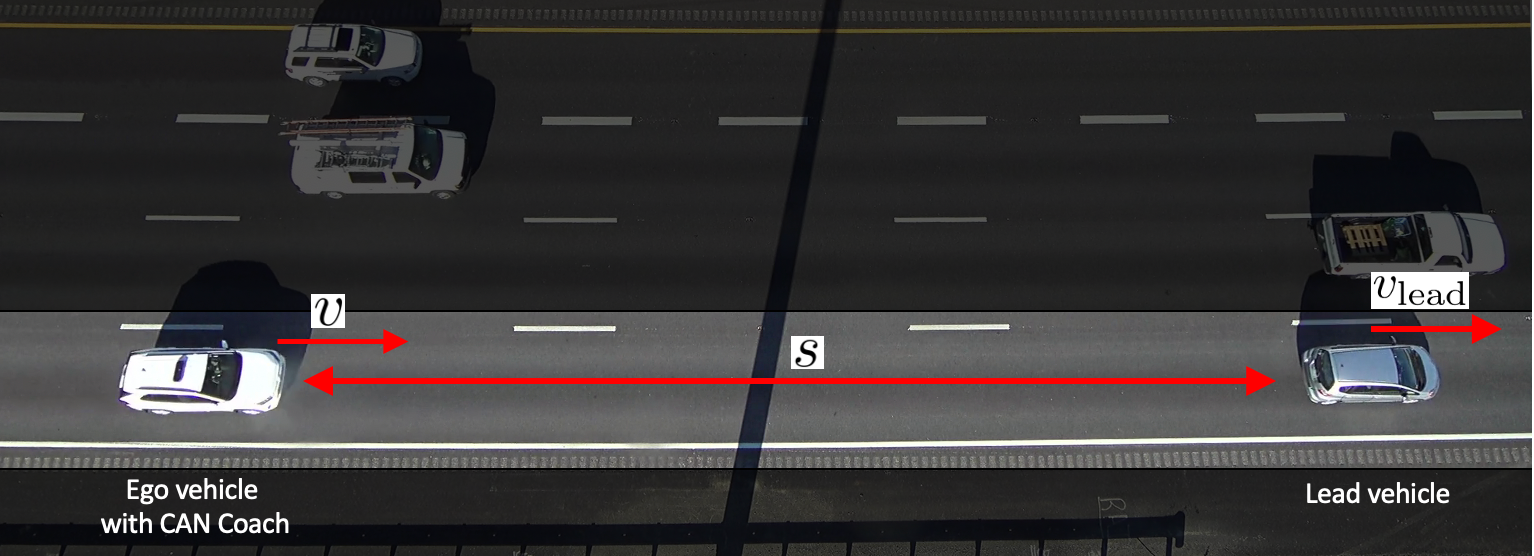}
  \caption{Ego vehicle with CAN Coach (White Toyota RAV4) and lead vehicle (Silver Honda Fit) in formation in the right lane.}
  \label{fig:teaser}
\end{teaserfigure}

\maketitle
\section{Introduction}

The goal of Lagrangian control is to control a few particles within a flow in order to effect the overall flow. In the domain of transportation, this may be carried out by control of a few vehicles in the flow of traffic in order to achieve a control objective. 
In order to regulate certain kinds of traffic phenomena, it is necessary to go beyond velocity control (which might leave space that encourages lane-changing) in order to control the spacing between the control (ego) vehicle and the lead vehicle. 

Research has been done to evaluate the idea of Lagrangian control of traffic with automated vehicles, demonstrating that vehicle automation can smooth traffic waves both in simulation~\cite{vinitsky2018lagrangian} and in full-scale experiments~\cite{stern2017dissipation}. These controllers dampen traffic waves through prescribed time-gap following, but they depend on advanced sensing and computation from the vehicles of tomorrow in order to close the loop with an accuracy and dependability that is on par with human-driven vehicles, with a requirement that 5\% or more of the vehicles must be controlled vehicles.

This raises a motivating question: can a human driver drive in a manner that dampens traffic waves, and thus accelerate the potential to adopt Lagrangian control by having the human perform some tasks that might otherwise require advanced sensors? If a human driver can be enabled to accurately follow commands that would enable Lagrangian control, then it may be possible to rapidly deploy technology in the vehicles of today that combines the advanced perception found in human drivers with advice from a supervisory controller that provides updated time-gap set points to carry out a control objective.

In this paper, we ask drivers to follow a vehicle (the lead vehicle) in their vehicle (the ego vehicle) in order to determine how to positively improve time-gap following through the use of a CAN Coach. This CAN Coach uses information from the CAN bus such as instantaneous velocity along with estimates of the relative position and velocity of the lead vehicle. These data were collected using only data from the CAN in the ego vehicle.\footnote{This work was conducted under IRB approval \#200343.} Drivers are asked to follow a lead vehicle with several different objectives, as well as to follow an imaginary vehicle (a ghost vehicle) by using the CAN Coach to follow a time-gap (a measure of time separation between the vehicles), but without the visual feedback to the human-in-the-loop from the lead vehicle. 

Our results show that when coached by the CAN Coach, human-in-the-loop drivers can closely control the time-gap between the lead vehicle and the ego vehicle, i.e. space as a function of speed. The driving behavior with the CAN Coach is significantly improved compared to when subjects are instructed to use their own perception to achieve the control objective. When following the ghost vehicle, performance degrades due to the lack of a human-perceivable vehicle. 
The contribution of this paper is the demonstration of an HCPS that is effective in controlling an individual vehicle which could be used for Lagrangian control of traffic flow. The CAN-based feedback system that is used (CAN Coach) enables drivers to achieve precise control objectives they could not otherwise achieve by giving extrasensory feedback informing adjustments of velocity and spacing. In other words, the CAN Coach can coach drivers to achieve ``superhuman'' driving tasks, and enables a human-in-the-loop to drive in a manner that could dampen traffic waves.\footnote{Here, we refer to superhuman in a colloquial sense, i.e., improving on what human drivers can do without assistance.} If successful, the application could scale widely given the low cost of the technology and the pervasive availability of the vehicle sensors; this would demonstrate the feasibility to deploy Lagrangian control techniques with the vehicle fleet of today.

The remainder of this article is organized as follows. In Section~\ref{sec:related_work}, we discuss works informing the formation of CAN Coach. In Section~\ref{sec:can_coach_system_background}, we provide an overview of CAN data, the hardware platform, and the software platform. In Section~\ref{sec:methodology}, we describe the feedback design and the experimental design. Section~\ref{sec:results} describes CAN data validation, and results of the CAN Coach system experiments. Finally, Section~\ref{sec:conclusion} outlines the planned extensions from our work presented here.

\section{Related Work} 
\label{sec:related_work}


To effect traffic phenomena such as traffic waves via Lagrangian control, it is necessary to be able to control the velocity and spacing of individual vehicles. One approach to reduce the severity of traffic jams is through vehicular-based~\cite{forster2014cooperative} variable speed limits, which is an adaptation of classical infrastructure based approaches~\cite{breton2002shock}. The work \cite{delle2019autonomous} considers models for autonomous vehicles to control traffic flow composed primarily of human-piloted vehicles.

Designing human-in-the-loop vehicle control systems via modeling and simulation~\cite{delle2019autonomous,breton2002shock,forster2014cooperative,sullman2015eco} has drawbacks due to the fidelity of the modeling.  Driver assistance systems to promote truck eco-driving at signalized intersections have been tested on real vehicles in~\cite{wang2019early}. Similarly, field experiments demonstrating human traffic flow stabilization using an automated vehicle are considered in~\cite{stern2017dissipation}. 

The use of CAN data as a data source for analysis of driving behavior such as lane changing, turning, and driver categorization is    \cite{fugiglando2018driving,wakita2006driver} becoming increasingly recognized. In the work \cite{carmona2015data},  a driver behavior identification tool is presented that fuses CAN data with sensor data from an inertial measurement unit and a GPS unit to classify normal and aggressive maneuvers in real-time. Unlike CAN Coach introduced next, the analysis is not used to close the loop on the human driver in real-time.  However it does establish the potential of using sensor fusion at run-time to understand driver behavior. 

Using technology to improve vehicle control and properties of traffic is a longstanding area of interest. Notable examples include the use of \textit{Cooperative Adaptive Cruise Control} (CACC) \cite{milanes2013cooperative,naus2010string} systems, which provide longitudinal control of the vehicle using information exchanged between vehicles via vehicle-to-vehicle communication to improve stability and throughput. In this present work we take a different approach, and instead explore how sensing can enhance human drivers in a control task, rather than automating the task and relegating the human to a monitoring role.

Though there can be complex trade-offs in augmenting human driver sensing via ADAS, it can improve safety and comfort \cite{li2011cognitive}. This work does not aim to mimic current ACC with a human-in-the-loop; instead it explores how current vehicle technologies may be used to enhance human driving.

A closely related work to our own is~ \cite{wu2016traffic}, which implements a real-time speed advisory system using vehicle-to-vehicle communications to allow a driver to accurately match the speed of the vehicle ahead. As we show in Section~\ref{sec:results}, constant time-gap based feedback can offer the ability to regulate the vehicle speed and position, which is not possible using only velocity-based feedback. Moreover, using CAN data, we avoid the need for a vehicle-to-vehicle communication system, since radar-based relative speed and space-gap data is directly available on the CAN. 

\section{CAN Coach System Background}
\label{sec:can_coach_system_background}

This section provides an overview of the CAN data, hardware platforms, and the software platforms used in our study. 

\subsection{CAN Data}
To use the CAN Coach, depending on the control objective, we need measurements of the velocity of the ego vehicle, $v$, velocity of the leading vehicle, $v_{\mathrm{lead}}$ and the space-gap, $s$, the distance between the front bumper of the ego vehicle and the rear bumper of the lead vehicle (see Figure~\ref{fig:teaser}). By accessing the CAN data in the ego (following) vehicle, this information can be found via radar sensors and wheel encoders. 

Vehicles with ACC typically have front-facing radar, which directly measures information we need to compute the space-gap and relative velocity $v_{\mathrm{lead}}-v$. Note that for feedback in real-time, there are two related challenges with radar data. First, incoming radar data tracks many objects in a wide field of view and thus needs to be processed to extract a high quality estimate of the location and relative velocity of the lead vehicle. Second, any processing on the radar data must be done quickly so that the feedback to the human-in-the-loop remains relevant. These challenges---%
briefly discussed in Section~\ref{sec:cancoachnode}%
---are otherwise outside the scope of this paper.

\subsection{Hardware}
The base vehicle used in this work is a stock 2020 Toyota Rav4 Hybrid vehicle. The vehicle has as standard equipment an \textit{Adaptive Cruise Control} (ACC) system and \textit{Lane Tracing Assist} (LTA) system. The ACC system depends on a stock forward looking radar unit that transmits relevant data on the CAN. To access this data, we used a Gray Panda manufactured by Comma.ai~\cite{santana2016learning} as a data logging device. The Gray Panda is connected to a Raspberry Pi 4 via USB. The Pi decodes the messages, transforms them into Robotic Operating System (ROS) messages, executes the CAN Coach to generate an auditory feedback for the human-in-the-loop, and records all data from the car and from all ROS nodes.
 
\subsection{Software}
\begin{figure}
     \centering
     \includegraphics[width=0.75\columnwidth]{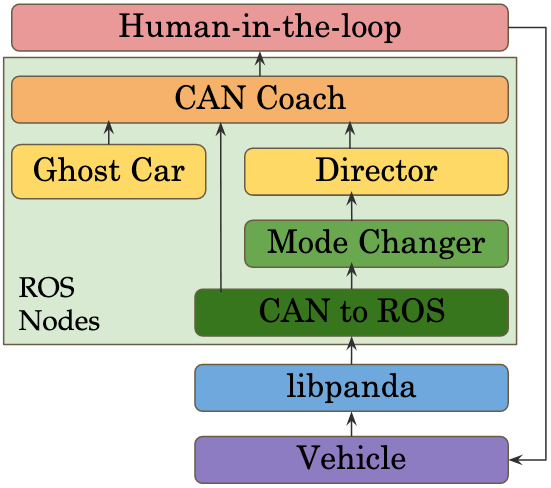}
     \caption{Diagram showing how information flows through the system.}
     \label{fig:stack}
 \end{figure}

    In this section, we discuss software that was developed to implement the CAN Coach System, execute the driving scenarios in a repeatable manner, and perform analysis of the captured data.  Figure~\ref{fig:stack} shows how the information flows through the system.


Comma.ai Panda devices are used as interfaces to allow for reading directly from the CAN bus.
Libpanda is a C++ based library to aid in the construction of custom software, abstracting the USB interface for easy reading and writing to the CAN bus and for easy reading of the GPS module (if installed) \cite{libpanda2020}. The libpanda library was developed in order to capture all data available from the connected CAN buses, rather than the subset of data selected by Comma.ai for use by their open-source control packages. 

The Robotic Operating System (ROS) is an open source framework for robotics. It provides the necessary tools and libraries to create and run peer-to-peer processes called nodes. ROS enables modular software development by utilizing reusable code packages~\cite{quigley2009ros}. 



The CAN Coach System consists of these ROS nodes:
\begin{itemize}
    \item CAN Coach node: the core node in the CAN Coach System, it processes relative vehicle velocity and distance and generates feedback for the human-in-the-loop;
    \item Ghost Mode: uses a simulated velocity of a lead vehicle in order to create a virtual vehicle's position in front of the ego vehicle;
    \item Director: facilitates communication between the \newline human-in-the-loop, mode changer, and CAN Coach, and ensures consistent execution of the experiment for each subject \textit{without} the need to have an additional researcher advancing each test mode;
    \item Mode Changer: allows the human-in-the-loop to advance (or reverse) the mode of the test while driving.
\end{itemize}




\subsubsection{CAN Coach}
\label{sec:cancoachnode}
 The CAN Coach, the eponymous ROS node, is what puts the human into the loop with the vehicle sensors, with audio feedback derived from CAN data in real time. The CAN Coach subscribes to velocity, relative distance, and relative velocity data obtained from the CAN bus, and when in the Ghost Mode subscribes to that node for relative distance and velocity of the virtual lead car. 
 
 With this information, the CAN Coach uses the 16 hi-frequency radar signals (20 Hz) that correspond to all tracked objects, along with a low-frequency radar trace (1 Hz) that corresponds to the lead vehicle's distance, in order to determine the relative distance $s$ and relative velocity $v_{\mathrm{lead}}-v$  between the ego and lead vehicles. This is achieved by finding a match between the most recent location of the lead vehicle and a buffer of recent raw radar data. With a match to the raw radar position, there is a corresponding relative velocity measurement. The CAN Coach can then communicate this information to the human-in-the-loop by producing the sound apt for the current control objective and feedback type.
 

\subsubsection{Ghost Mode}
\label{sec:ghost_mode}
The Ghost Mode is designed to measure how well the human-in-the-loop can track a lead vehicle through feedback from the CAN Coach, but without the ability to see the lead vehicle. Additional information exploring the motivation and merit of the Ghost Mode is provided in Section~\ref{sec:experimental_conditions}.

In order to carry out this mode of the experiment, relative distance and relative velocity are measured from the ego vehicle to a virtual `ghost' vehicle---rather than obtained from the CAN bus. Through code generated from a Simulink model, the ego vehicle's velocity signal and a simulated constant velocity of the ghost vehicle are integrated to simulate the relative distance and relative velocity between the ego vehicle and lead (ghost) vehicle. 

The CAN Coach calculates the time-gap in Ghost Mode from the ego velocity, and the difference in the distance traveled for both vehicles during ghost mode plus a constant of 65 m. A companion Stateflow model periodically determines whether the virtual space gap between the ego and ghost vehicles has either grown too large (which may result in unsafe high speed by the human-in-the-loop to catch up) or too small (which may result in unsafe low speed to allow the ghost vehicle to pass). If the virtual space gap $s > 100$ or $s<-30$, then the virtual distance is reset to 0~m (plus the constant 65~m) to permit safe execution of the experiment. The constant is added because at the initialized speed, 29 m/s, and time-gap set point, 2.25 s, the target space-gap is 65 m. This ensures that in the first minutes of Ghost Mode, if the ghost vehicle is reset it will return to the target state. 



\subsubsection{Director and Mode Changer Nodes}
\label{sec:director}
Due to ongoing health and safety practices, only one person is allowed in a vehicle at a time---which introduces challenges in carrying out driving experiments where a passenger might otherwise control the experiment mode. We address this challenge through these supervisory nodes that require no synchronous interaction from the driver: a state-based Director, and a Mode Changer.

\begin{figure}
\centering
\includegraphics[width=1.0\linewidth]{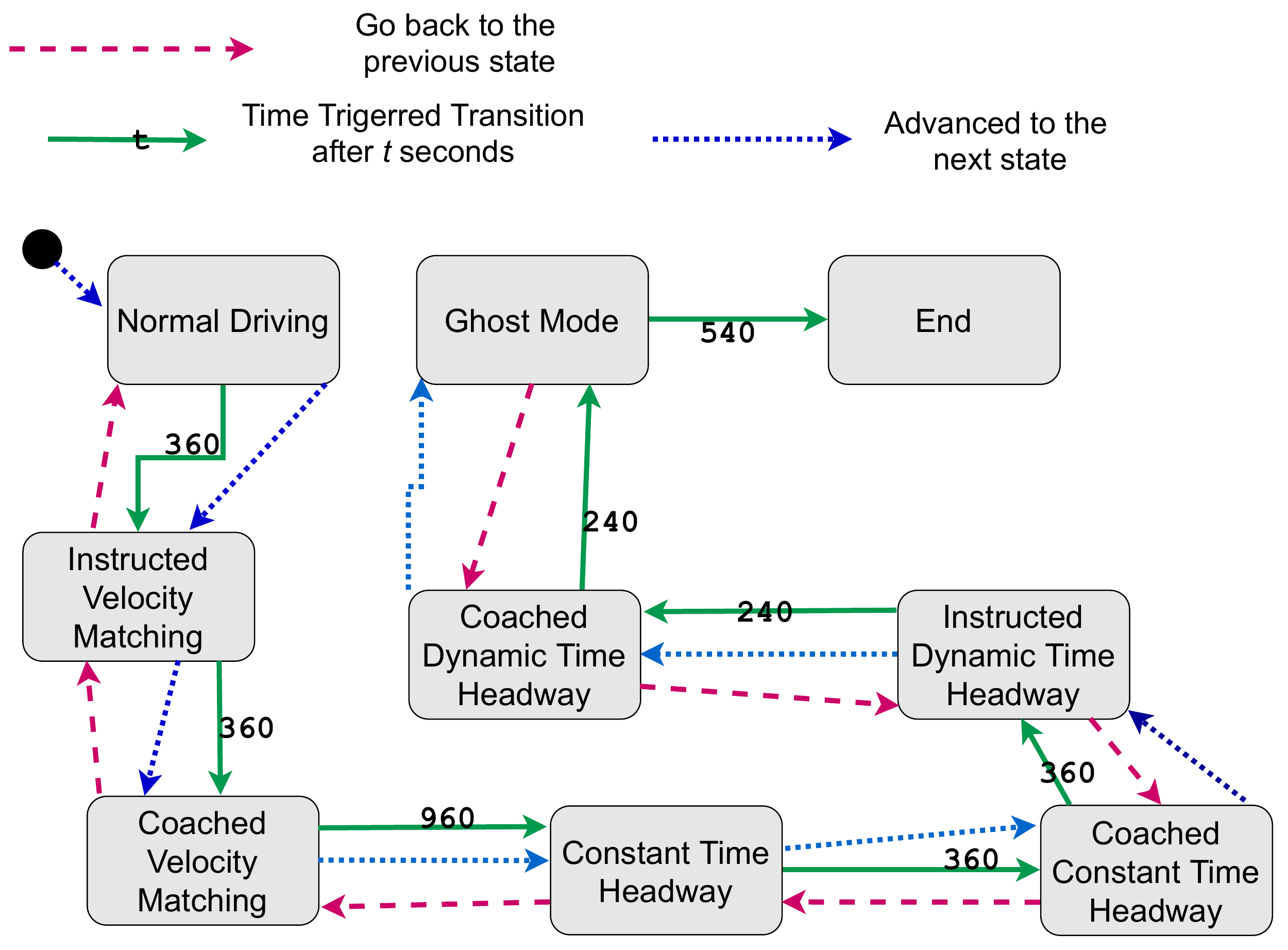}
\caption{The Director node: Each experiment state lasts for a precise duration, or can be advanced or reversed based on an input.
}
\label{fig:director_node}
\end{figure}

An abridged schematic of the Director node, simplified for readability, is provided in Figure~\ref{fig:director_node}. The Director advances modes based on elapsed time, yielding a precise, repeatable sequence with a fixed duration for each segment of the CAN Coach experiment. 
The Director node publishes the set point, feedback type, and current mode to the CAN Coach node every 0.5~s.

The Director subscribes to the Mode Changer, allowing the human-in-the-loop to advance (or reverse) the experiment mode. For example, if the lead driver thinks that the data would not be useful for analysis, that driver could request the human-in-the-loop subject to reset the mode using the vehicle's light stalk. The experiment's modes are the tested pairs of control objectives and feedback types.  

\section{Experimental Methods}
 \label{sec:methodology}
This section describes the decisions made to design the feedback within the CAN Coach, and to design the experiments.
\subsection{Feedback Design}
  The CAN Coach provides simple feedback to the human-in-the-loop, because complexity can be a burden~\cite{bazilinskyy2018graded}. In CAN Coach, when the human-in-the-loop is a threshold amount away from the control target, an audible signal is output to inform the human-in-the-loop of an action to take. We choose audio feedback because \cite{scott2008comparison} shows auditory driving warnings correspond with better driver reaction times than visual warnings. Some commercial systems use discrete audible feedback, such as Forward Collision Warning Systems. They take into account both the physical driving context and human interaction; the precise timing of the discrete intervention is key to the system's success \cite{jamson2008potential, abe2006alarm}. Literature suggests, however, that continuous feedback \cite{seppelt2007making,fairclough1997effect} is preferred over discrete feedback \cite{risto2014supporting} for CAN Coach, because continuous feedback allows for greater control of precision over an extended period of time.
  
  Under time-gap control, if the current time-gap differs from the desired time-gap (the set point) by more than $\pm$0.05~s, a sound is output to increase or shorten the time-gap. Under velocity matching control, if the relative velocity to the vehicle ahead differs by more than $\pm$0.4 m/s, a sound is given as output to increase or decrease the velocity.  
  
  There are only three messages the feedback communicates: speed up, slow down, or do nothing. A high pitched sound is used to indicate the human-in-the-loop should accelerate; a low pitched sound is used to indicate the human-in-the-loop should decelerate. No sound is emitted when the human-in-the-loop is near the set point. 
    
\subsection{Experimental Design}
 
  \subsubsection{Description of the Driving Environment} 
\begin{figure}
        \centering
        \includegraphics[trim={0.0cm 0.0cm 0.0cm 1.3cm},clip,width=1.0\linewidth]{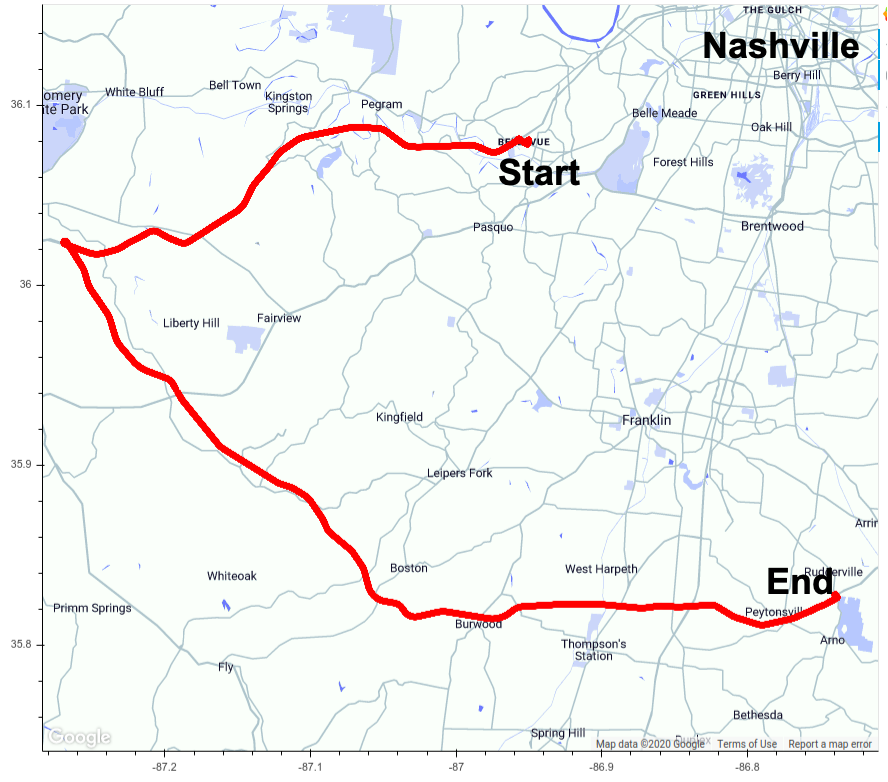}
        \caption{Route used to conduct the CAN Coach experiments.}
        \label{fig:route}
    \end{figure}

    The \newline experiments conducted in this work are conducted with two vehicles on a freeway. A lead vehicle is used during the experiment to provide consistency in the testing environment, and to trace the route for the ego vehicle, at a speed of 65 mph ($\approx 29.0\;\mathrm{m/s})$. The ego vehicle is instrumented with the CAN Coach, and it follows immediately behind the lead vehicle while executing desired control objectives. Each drive begins and ends in the same locations for each tested driver, for a consistent roadway grade along the route on which the drivers are tested. 
    The tests occurred on a 55 mile freeway route (Figure~\ref{fig:route}) at off peak hours to reduce interactions between the vehicles involved in the experiment and other road users. Each experiment lasts about 57 minutes per driver.
    
   Prior to the start of each experiment, each driver is given an overview of the experiments, the route, the technologies used in the experiments, and the safety protocols. Drivers are informed to follow instructions given during the experiment to the best of their ability without compromising safety. Additional instructions are provided to the driver automatically using a standardized script prompted from the Director (Section~\ref{sec:director}), eliminating the need for any experimental staff to be in the vehicle with the human-in-the-loop subject.
   


At the start of the experiment, each driver is given a period of time to operate the vehicle following the lead vehicle under a \textit{Normal Driving} regime. The driver is given instructions to follow the vehicle ahead without changing lanes.  

Recall that the lead vehicle velocity is denoted by $v_\text{lead}$ and the ego vehicle velocity be denoted by $v$ (Figure~\ref{fig:teaser}). The relative velocity is $\Delta v:= v_\text{lead}-v$, where a %
negative 
relative velocity indicates the ego vehicle is catching up, while a %
positive 
relative velocity indicates the ego vehicle is falling behind. The \textit{space-gap}, denoted by $s$, is the distance between the front bumper of the ego vehicle and the rear bumper of the lead vehicle. The \textit{time-gap}, denoted $\tau:=s/v$, is the time required to travel the space-gap distance $s$ when traveling at the velocity $v$ \cite{loulizi2019steady}.

In all experiments, the lead vehicle speed is set at 29.0 m/s (65 mph) using a standard cruise control system. Variations in the lead vehicle speed occur due to variations in the grade of the roadway as well as traffic conditions.

Six drivers consisting of university students and staff were recruited to conduct the experiments. All drivers passed a university-required driver safety course. 

\subsubsection{Description of Control Objectives}
    \label{sec:experiment_protocol}
During each experiment, the human-in-the-loop driver of the ego vehicle is asked to adjust the speed of the vehicle to achieve a desired control objective. Only one objective is used at a time. The control objectives are as follows. 
     
     \begin{itemize}
         \item \textit{Constant Time-gap}:
        Drive the ego vehicle such that the time-gap to the vehicle ahead is $\tau=2.25$ s.
         \item \textit{Velocity Matching}:
        Match the velocity of the lead vehicle (i.e. so that $\Delta v=0$). The set speed of the lead vehicle is not announced to the driver (i.e., they are not informed that the lead vehicle is traveling at a fixed speed of 29.0 m/s). This meaningfully differs from Constant Time-gap because there is no fixed target for following distance.
         \item \textit{Dynamic Time-gap:}
        Drive the ego vehicle such that the time-gap is $\tau=2.25$~s or $\tau=1.8$~s. The desired time-gap changes every 60~s.
         
         
     \end{itemize} 
 
 \subsubsection{Feedback Type}
  \label{sec:experimental_conditions}
 To meet the control objective, the driver is given varying degrees of information:  
    \begin{itemize}
        \item \textit{Instructed Driving (human visual perception feedback)}: The human-in-the-loop is verbally given the specific control objective (e.g., drive with a specified time-gap). In the case of the dynamic time-gap control objective, the human-in-the-loop is told when the time-gap changes, and the value of the new time-gap. 
        \item \textit{Coached Driving (human visual perception and CAN feedback)}: The human-in-the-loop is given all of the information provided in the Instructed Driving setup, as well as additional feedback using CAN data. 
        \item \textit{Ghost Mode (CAN feedback but without human visual perception)}: Ghost mode is a special setup in which the lead vehicle is replaced by a virtual (i.e., ghost) vehicle. CAN Coach is used to provide feedback to help the human-in-the-loop of the ego vehicle achieve a control objective relative to the ghost vehicle. 
        The point of the ghost mode is to remove the possibility of the human-in-the-loop from visualizing the space-gap or the velocity of the lead vehicle. The human-in-the-loop is given the same instructions and feedback as for Coached Driving, with the major modification that the real lead vehicle is replaced by the ghost vehicle.
    \end{itemize}

\section{Results}
\label{sec:results}
This section presents the results from each human-in-the-loop subject under different control objectives and feedback types.  We first consider the constant time-gap control objective, and compare the performance of humans-in-the-loop using Instructed Driving to the performance when CAN Coach is used to provide feedback. We show that CAN Coached driving improves the ability of the human-in-the-loop to achieve the control objective. Next, we consider the importance of the the ability of the human-in-the-loop to see the vehicle ahead by comparing driving under Coached Driving, to driving in Ghost Mode.  Finally we consider additional control objectives including velocity matching and dynamic time-gap matching.

\subsection{Data Preprocessing}
\label{sec:data_preprocessing} CAN velocity and space-gap data are validated using GPS devices following the process described in~\cite{wang2019estimating}. Irrelevant data from the experiments must be removed in preprocessing. These data are collected due to experimental conditions on an open roadway. For example, we discard data near and on the ramp connecting the two freeways that form the driving route. For some drivers during some tests, a vehicle not part of the experiment can have undue influence the experiment, e.g., by cutting in between the lead vehicle and the instrumented ego vehicle. These data are removed prior to analysis. Through manual inspection of dash-mounted video data recorded during the tests, we determine simple thresholds that remove data corresponding to these anomalies outside of the intended testing environment. Specifically, we observe that we can eliminate data corresponding to these events by discarding data below the 10th percentile velocity, and when the relative velocity is below the 5th percentile or above the 99th percentile. 

\subsection{Constant Time-gap Control}
    In the first set of experiments, we compare the ability of a human-in-the-loop to achieve the constant time-gap control objective. We first present the results under Instructed Driving (visual feedback) compared to Coached Driving (visual and CAN feedback) conditions. Then we remove the possibility of visual feedback by considering a constant time-gap control to a ghost vehicle.
    
    We consider the following performance measures. The control objective is to achieve a desired time-gap of $\tau_\text{desired}= 2.25$~s. The time-gap error at each measurement time instant $t$ is computed as
        $\varepsilon_\tau(t):=\tau_\text{desired}-\tau(t)$.
     The empirical distribution of the time-gap error over the duration of the test can be computed, as well as the mean error and the standard deviation of the error. Similarly, we can compute a corresponding space-gap error as $\varepsilon_s(t)=v(t)\tau_\text{desired}-s(t)$, and we can compute the space-gap error statistics.


     \begin{figure}
     \centering
     \begin{subfigure}{\linewidth}
         \centering
         \includegraphics[width=\textwidth]{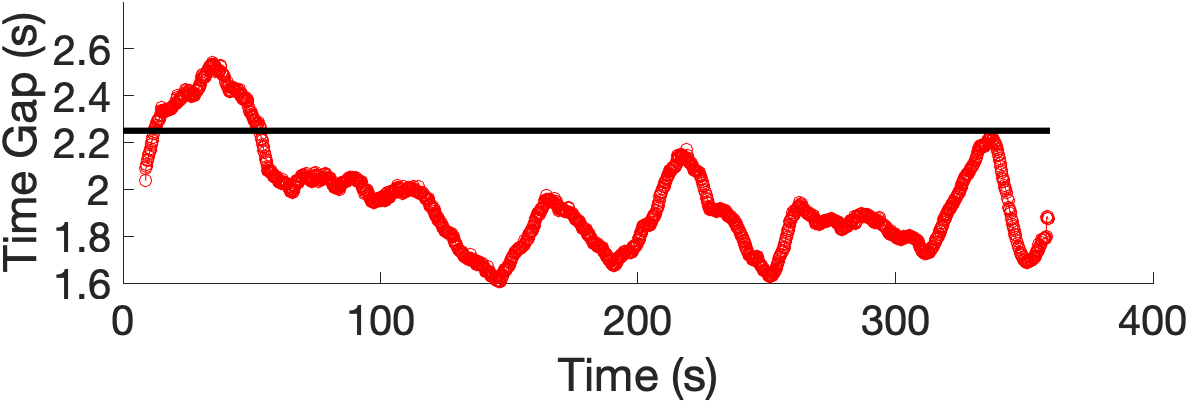}
         \caption{Instructed Driving} 
         \label{fig:timeseriesI}
     \end{subfigure}
     \begin{subfigure}{\linewidth}
         \centering
         \includegraphics[width=\textwidth]{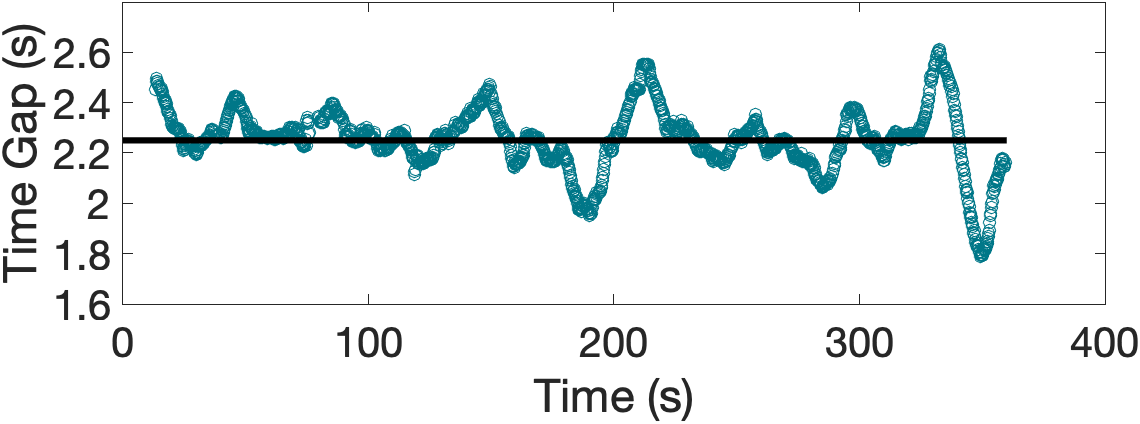}
         \caption{Coached Driving}
         \label{fig:timeseriesC}
     \end{subfigure}

        \caption{Driver 1. (a) Instructed Driving and (b) Coached Driving for the Constant Time-gap control objective. Horizontal line indicates the desired time-gap of 2.25 s.}
        \label{fig:timeseries}
    \end{figure}
    
    \subsubsection{Instructed vs. Coached Driving: Driver 1}
\begin{figure}
        \centering
        \includegraphics[width=\linewidth]{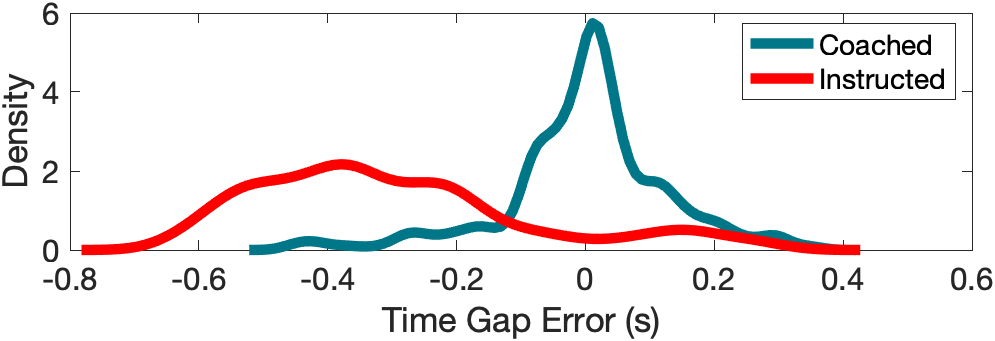}
        \caption{Driver 1 time-gap error distribution for Instructed and Coached Driving under Constant Time-gap control.}
        \label{fig:cthtgkde}
    \end{figure}

        We \newline compare~the results of a single driver under Instructed Driving to Coached Driving.  Under the Instructed Driving (where feedback is based only on human visual perception), Driver 1 does not achieve the control objective. Figure~\ref{fig:timeseriesI} shows 6 minutes (360 seconds) of instructed driving. The driver has a time-gap that is consistently too low. In contrast, Figure~\ref{fig:timeseriesC} shows the same driver and control objective under the Coached setting in which visual and CAN feedback are provided. It is clear from the figures that CAN Coach helps the driver correct the time-gap to achieve the control objective.
        To summarize the time-series shown in Figure~\ref{fig:timeseries},  histograms of the time-gap error with and without CAN feedback is provided in Figure~\ref{fig:cthtgkde}. Under Instructed Driving, the driver has a mean time-gap error of -0.3~s, and a standard deviation of 0.22~s. Under Coached driving, the mean time-gap error is 0.02~s, and the standard deviation is 0.13~s.  In summary, the addition of feedback based on the CAN reduces the mean time-gap error by 93\% and the standard deviation by 41\%.
        
\begin{figure}
         \centering
         \begin{subfigure}{0.49\linewidth}
             \centering
             \includegraphics[width=\textwidth]{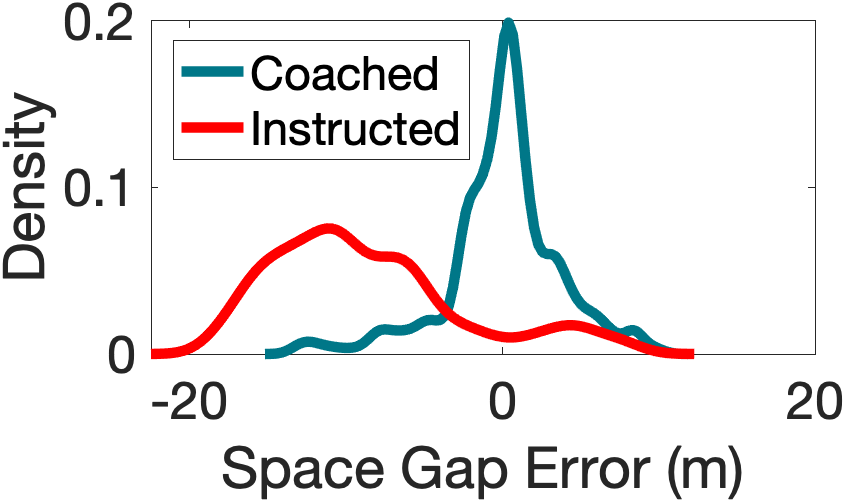}
             \caption{Space-gap Error Distribution}
             \label{fig:sgcomponent}
         \end{subfigure}
         \begin{subfigure}{0.5\linewidth}
             \centering
             \includegraphics[width=\textwidth]{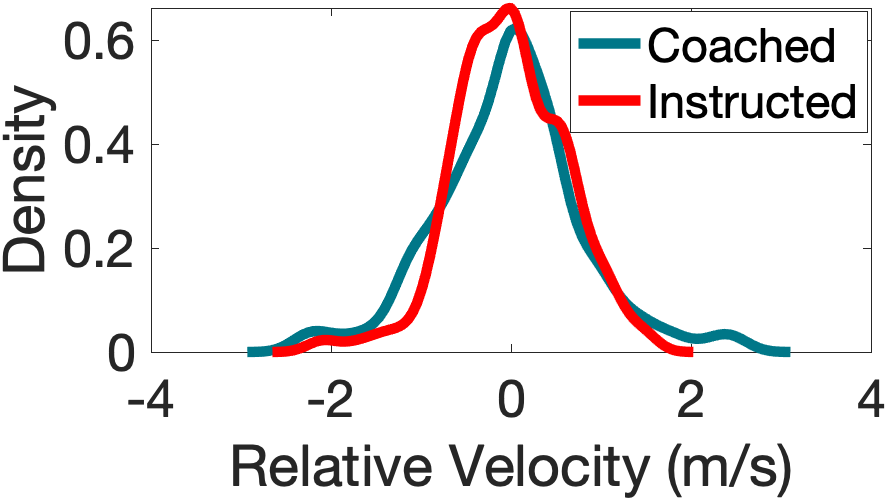}
             \caption{Relative Velocity Distribution}
             \label{fig:relvcomponent}
         \end{subfigure}
        \caption{Driver 1 space-gap error and relative velocity to the lead vehicle under Constant Time-gap control. }
        \label{fig:components}
    \end{figure}

      It is also possible to explore the corresponding space-gap error and the relative velocity to the lead vehicle under Instructed and Coached Driving. Comparing \ref{fig:sgcomponent} and \ref{fig:relvcomponent}, we see that though there are marginal differences in relative velocity between Instructed and Coached Driving, there are important differences in the space-gap error. Coached driving substantially reduces the mean space-gap error (from -8.71~m to 0.088~m) as well as the standard deviation of the space-gap error (from 6.15~m to 3.65~m).
    
    \subsubsection{Instructed vs. Coached Driving: All Drivers}
\begin{figure}
        \begin{subfigure}{\linewidth}
         \centering
         \includegraphics[trim={0.0cm 0.0cm 0.0cm 0.0cm},clip,width=\textwidth]{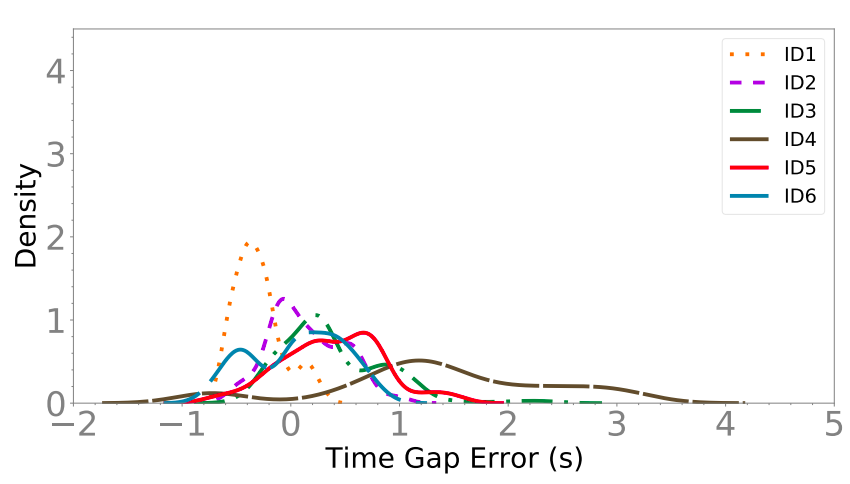}
         \caption{Instructed Driving}
        \label{fig:alldriverI}
        \end{subfigure}
    \begin{subfigure}{\linewidth}
        \centering
        \includegraphics[trim={0.0cm 0.0cm 0.0cm 0.0cm},clip,width=\textwidth]{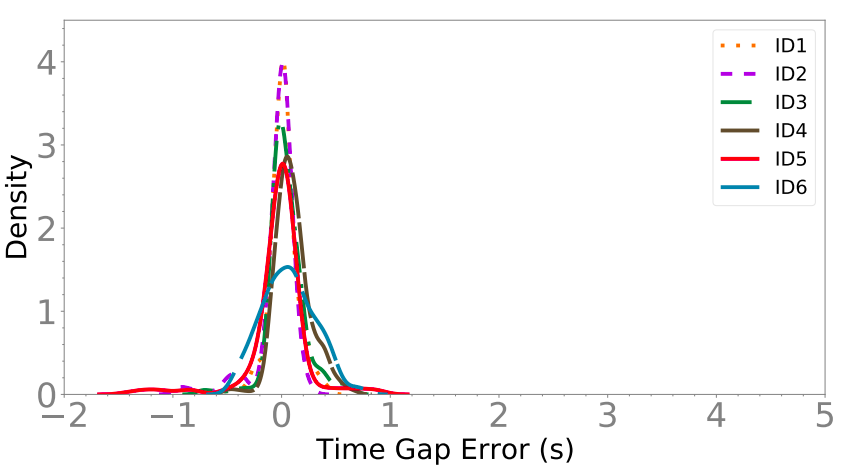}
        \caption{Coached Driving}
        \label{fig:alldriverC}
        \end{subfigure}
        \caption{Time-gap error distributions for all drivers under (a) Instructed Driving and (b) Coached Driving under Constant Time-gap control.}
        \label{fig:allDrivers}
    \end{figure}

        Figure \ref{fig:allDrivers} shows the results comparing instructed driving to coached driving for all six drivers. Examining Figure~\ref{fig:alldriverI}, it is clear that the drivers had wide performance variation under Instructed Driving. This is true both in terms of average error from the set point, as well as the standard deviation. While aiming for a 2.25~s time-gap, the observed average time-gaps range from from 1.95 s (for Driver 1) to 3.61 s (for Driver 4). The standard deviation of the time-gap range from 0.22~s (Driver 1) to 1.01~s (Driver 4).  

        Figure~\ref{fig:alldriverC} presents the time-gap errors for all six drivers under Coached Driving. All drivers are able to obtain a low mean error on the time-gap. The smallest time-gap error is 0.02~s (Driver 1) and the largest is 0.1~s (Driver 4). The variance of the errors are also substantially reduced. The standard deviations range from 0.13~s (Driver 1) to 0.28~s (Driver 5). To put the standard deviations into perspective, note that the standard deviations of all six drivers under Coached Driving are lower than the second most consistent driver (Driver 2; standard deviation of 0.34~s) under the Instructed mode.
        
        Across all drivers (Table~\ref{tab:CTHsummary}), the mean time-gap error reduced from 0.44~s under Instructed Driving to 0.05 s under Coached Driving, which is a 73\% reduction in mean error. The drivers on average are more consistent, with the error standard deviation reduced from 0.49~s to 0.2~s (a 53\% reduction). Only Driver 6 saw no reduction in mean error when operating under coached mode, which is due to the fact that Driver 6 had a very low 0.06~s mean error in instructed mode. All other drivers saw mean error reductions above 70\% as a result of the feedback from the CAN bus.

               \begin{table}
\begin{tabular}{@{}ccccccc@{}}
\toprule
             & \multicolumn{2}{c}{Instructed} & \multicolumn{2}{c}{Coached}  & \multicolumn{2}{c}{\% reduction} \\ \midrule
Driver       & mean           & std    & mean          & std    & mean              & std        \\\midrule
1            & 0.30            & 0.22          & 0.02          & 0.13         & 93\%              & 41\%              \\
2            & 0.14           & 0.34          & 0.04          & 0.18         & 71\%              & 47\%              \\
3            & 0.38           & 0.48          & 0.04          & 0.18         & 89\%              & 63\%              \\
4            & 1.36           & 1.01          & 0.10           & 0.16         & 93\%              & 84\%              \\
5            & 0.39           & 0.46          & 0.03          & 0.28         & 92\%              & 39\%              \\
6            & 0.06           & 0.42          & 0.06          & 0.24         & 0\%               & 43\%              \\
\midrule
\textbf{Avg} & \textbf{0.44}  & \textbf{0.49} & \textbf{0.05} & \textbf{0.2} & \textbf{73\%}     & \textbf{53\%}     \\ \bottomrule
\end{tabular}
       \newline\caption{Performance summary of Instructed and Coached driving under Constant Time-gap control. Units in sec.; Percent reduction is computed using Instructed as the baseline. }
       \label{tab:CTHsummary}
\end{table}

\subsubsection{Removing Visual Feedback via Ghost Mode}
    In this section we consider the importance of human-based visual perception in the control loop.  The expectation is that by removing the ability of the human-in-the-loop to also sense the vehicle ahead, the ability to follow the vehicle will degrade. The use of Ghost Mode allows for an ablation analysis because it removes the human perception of the lead vehicle by replacing it with an abstract ghost vehicle. 
    
\begin{figure}
    \centering
    \includegraphics[width=\linewidth]{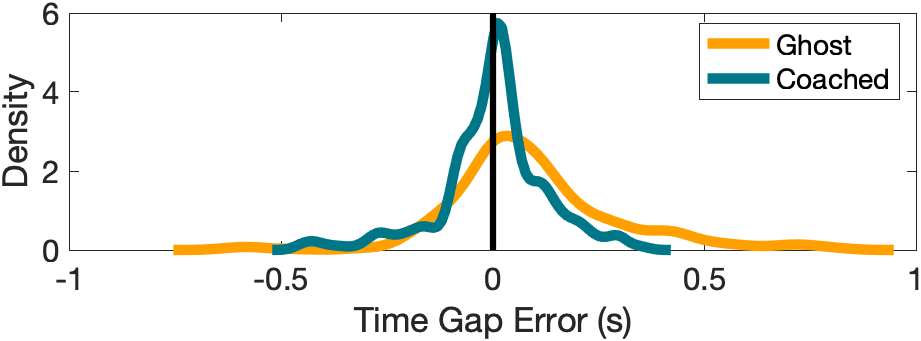}
    \caption{Driver 1. Time-gap error distribution for Coached Driving and Ghost Mode feedback under Constant Time-gap control.}
    \label{fig:ghostcthtg}
    \end{figure}

    We first highlight the results of Driver 1 and then present the summary results for the remaining drivers. Figure~\ref{fig:ghostcthtg} shows the time-gap error under Coached and Ghost Mode driving conditions. There is a drop in performance of the human-in-the-loop when following a ghost vehicle compared to a real vehicle, as quantified by the mean time-gap error. The mean time-gap error of 0.02~s under Coached Driving conditions increases to 0.15~s in Ghost Mode. Similarly, the standard deviation increases from 0.13~s in Coached Driving to 0.26~s when in Ghost Mode.
    
\begin{figure}
        \begin{subfigure}{\linewidth}
        \centering
        \includegraphics[width=\textwidth]{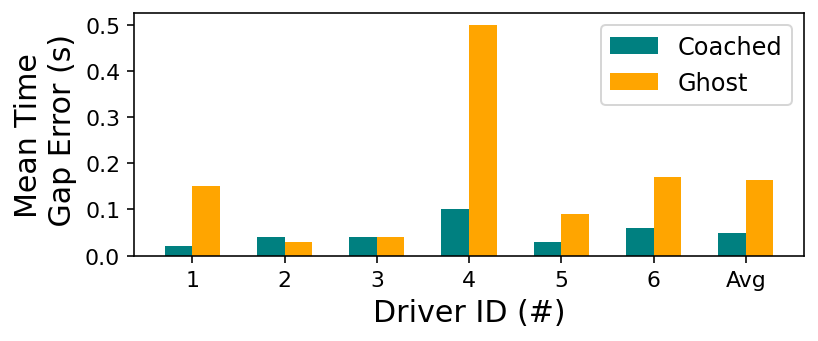}
        \caption{Mean time-gap error for Coached Driving and Ghost Mode.}
        \label{fig:meanDiff}
        \end{subfigure}
        
        \begin{subfigure}{\linewidth}
        \centering
        \includegraphics[width=\textwidth]{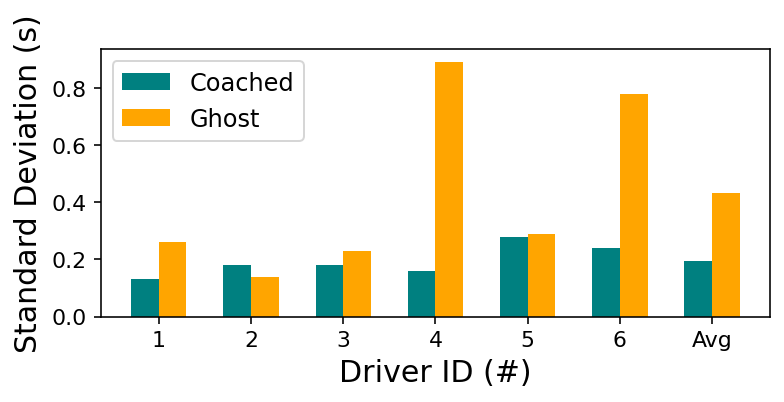}
        \caption{Standard deviation of the time-gap for Coached Driving and Ghost Mode.}
        \label{fig:stdDiff}
        \end{subfigure}

        \caption{Comparison of mean time-gap error and standard deviation for Coached Driving and Ghost Mode.}
        \label{fig:differences}
    \end{figure}

    Next we consider the range of performances across the drivers. Figure~\ref{fig:differences} summarizes the mean time-gap error and standard deviation of the time-gap error under Coached Driving and Ghost Mode driving. Figure~\ref{fig:meanDiff} shows the the mean time-gap error increases for Drivers 1, 4, 5, and 6 when following a virtual vehicle rather than a real one. The mean time-gap error is identical for Driver 3 under both feedback modes, while the mean error for driver 2 is slightly reduced under Ghost Mode relative to Coached Driving. The mean time-gap error across all drivers is 0.05~s during Coached Driving, compared to 0.13 s in Ghost Mode. The standard deviation of the time-gap error distributions under Coached and Ghost driving is shown in Figure~\ref{fig:stdDiff}. The standard deviation increases for all drivers except Driver 2 when following a ghost vehicle compared to a real one.  Combined, all drivers except Driver 2 saw either an increase in the mean error, standard deviation of the error, or both. This indicates that most of the humans-in-the-loop contribute to achieving the control task by seeing the vehicle ahead, even with CAN Coach feedback.
  
    Note that performance under Ghost Mode is still improved compared to driving using only visual perception in the Instructed mode.
    
\subsection{Velocity Matching Control}
    Next we consider a new control objective, in which the human-in-the-loop is asked to match the velocity of the vehicle ahead. There is a moderate reduction in standard deviation of the relative velocity for Coached Driving over Instructed Driving for the velocity matching control objective. Coached Driving feedback reduces mean relative velocity by 0.03 m/s, and reduces standard deviation by 0.16 m/s compared to Instructed Driving. The improvement is modest because the drivers under Instructed Driving were able to match the velocity of the lead vehicle reasonably well. As a consequence, CAN feedback resulted in only a slight reduction in error.

\begin{figure}
     \begin{subfigure}{\linewidth}
         \centering
\includegraphics[trim={0.0cm 0.0cm 0.0cm 0.0cm},clip,width=\linewidth]{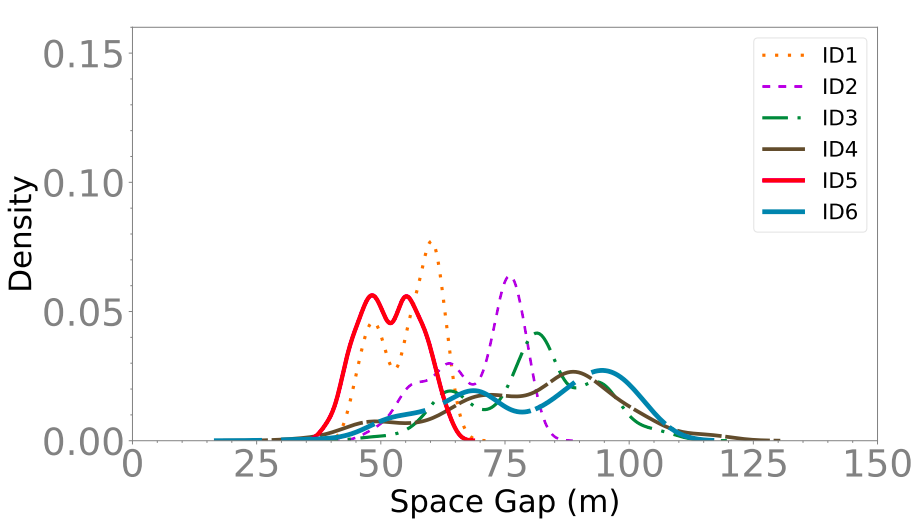}
    \caption{Velocity matching}
    \label{fig:sgvariation}
    \end{subfigure}
         \begin{subfigure}{\linewidth}
         \centering
\includegraphics[trim={0.0cm 0.0cm 0.0cm 0.0cm},clip,width=\linewidth]{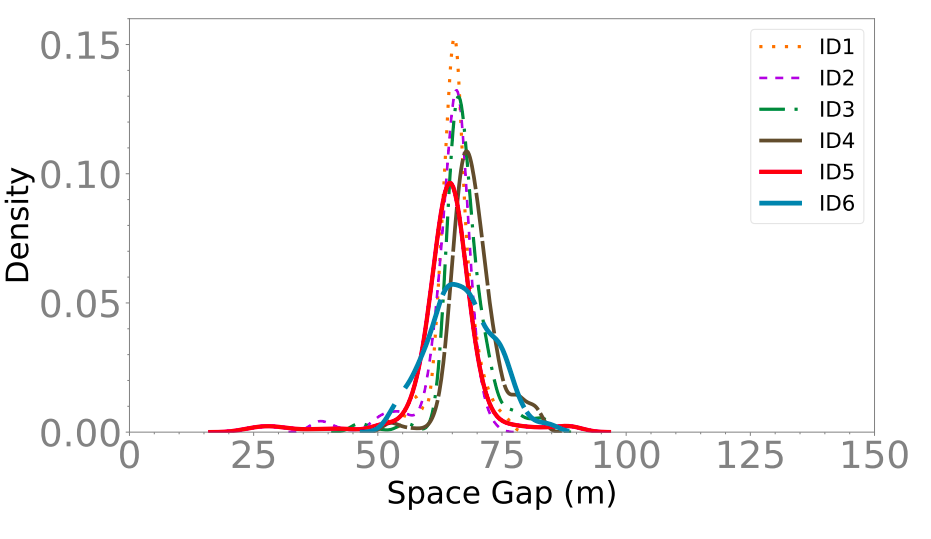}
    \caption{Constant Time-gap}
    \label{fig:sgvariation_coached_constant_th}
    \end{subfigure}
    \caption{Space-gap distributions with Coached feedback for (a) Velocity Matching and (b) Constant Time-gap control.}
    \end{figure}

The most important and intuitive observation is that Velocity Matching does not standardize the space-gap across the drivers. The space-gap distribution under Coached Velocity Matching is shown in Figure~\ref{fig:sgvariation}, which can be compared to the space-gap variation under the constant time-gap control objective (Figure~\ref{fig:sgvariation_coached_constant_th}). When under velocity matching control, mean space-gaps ranged from 51.8~m (Driver 5) to 80.5~m (Driver 3). Under time-gap control similar relative velocity errors are observed, but the mean space-gaps are more tightly bounded. They range from 64.0~m (Driver 2) to 69.2~m (Driver 4). In summary, controlling the time-gap provides similar velocity control and better space-gap regulation than velocity based control.  
    

    
\subsection{Dynamic Time-gap} 
\label{sec:dth}
\begin{figure}
        \begin{subfigure}{\linewidth}
        \centering
    \includegraphics[trim={0.0cm 0.0cm 0.0cm 0.0cm},clip,width=\textwidth]{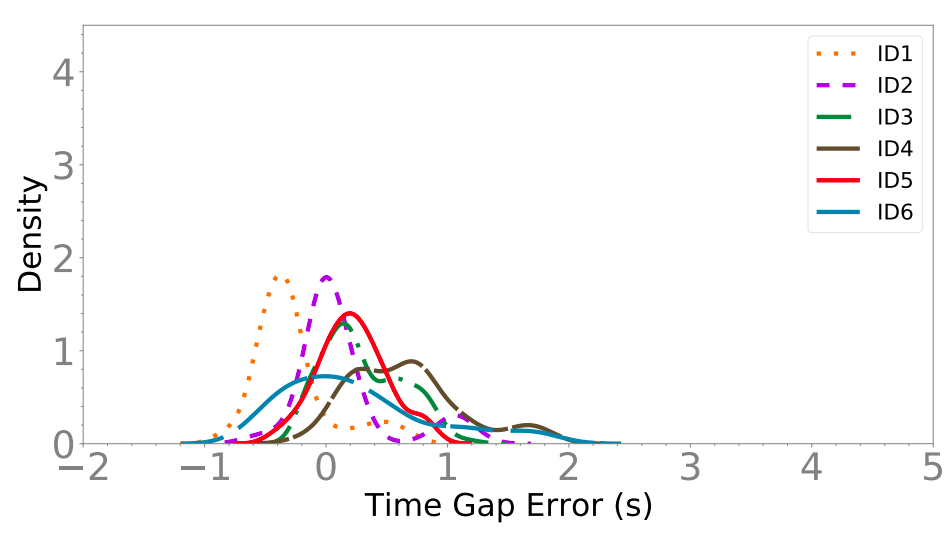}
        \caption{Instructed Driving}
        \label{fig:dthinstructed}
        \end{subfigure}
        \begin{subfigure}{\linewidth}
        \centering
     \includegraphics[trim={0.0cm 0.0cm 0.0cm 0.0cm},clip,width=\textwidth]{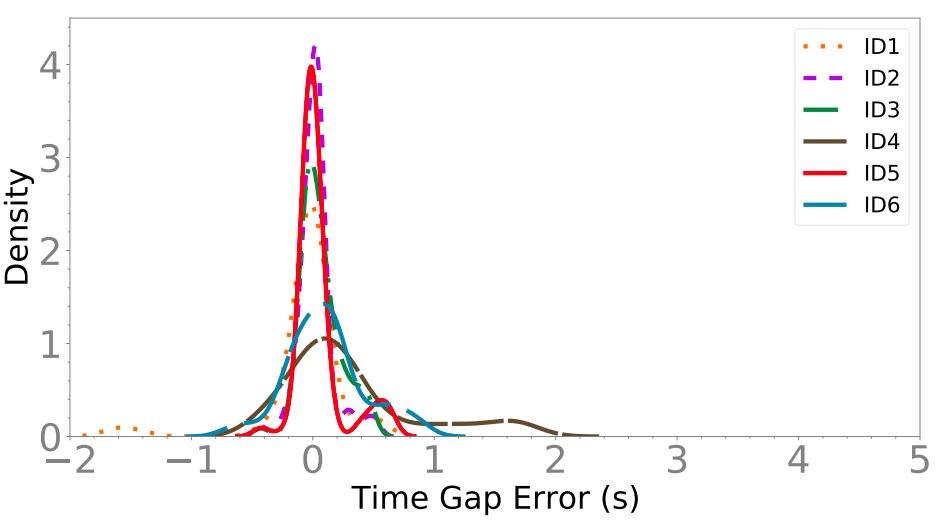}
        \caption{Coached Driving}
        \label{fig:dthccerror}
        \end{subfigure}
        \caption{Time-gap error distributions for all drivers under (a) Instructed Driving and (b) Coached Driving under Dynamic Time-gap control.}
        \label{fig:dth}
    \end{figure}

     Finally, we consider the consequences of changing the time gap between two fixed values in a Dynamic Time-gap objective.  Figure~\ref{fig:dthinstructed} shows the time-gap error distribution under the Instructed Driving conditions, while Figure~\ref{fig:dthccerror} shows the time-gap error distribution under the Coached Driving condition. The mean time-gap error decreases from 0.21 s (Instructed) to 0.08 s (Coached). The standard deviation is decreased from 0.40 s (Instructed) to 0.30 s (Coached). This shows that CAN-based feedback can more accurately and precisely control the vehicle than without CAN-based feedback, even as the control objective increases in difficulty. 
     
     When comparing Dynamic Time-gap Coached Driving to the Constant Time-gap Coached Driving, we see that performance was slightly worse with Dynamic Time-gap than with Constant Time-gap. From Constant Time-gap Coached Driving to Dynamic Time-gap Coached Driving, the mean time-gap error increased from 0.05 s to 0.08 s; the mean standard deviation increased from 0.19 s to 0.30 s.
    
\section{Conclusions and Future Work}
\label{sec:conclusion}
With an eye towards scalable Lagrangian traffic control, this work considered the possibility of vehicular control through a human-in-the loop system. By providing the driver feedback using sensor data reported on the CAN, drivers of varying skill levels are able to consistently achieve a desired time-gap control objective which effectively regulates both the velocity and the space-gap of the ego vehicle. The CAN Coach can coach drivers to achieve driving tasks that were not possible without feedback, opening a door to modify driver behavior for the benefit of the overall traffic flow.
Though changing time-gap affects safety and efficiency, this work focuses on the successful implementation of a control objective, not the merits of the control objective itself.  

In our future work we are interested in understanding the limits on the complexity of the control objective. For example it may be possible for drivers to explicitly follow a complex desired trajectory using time-gap based feedback to the desired trajectory, or for drivers to be coached to explicitly achieve a traffic wave stabilizing control policy. Enhanced driver assistance interfaces and experiments with a larger cohort of drivers will be important to generalize the findings presented here. Personalized feedback to drivers could improve performance and minimize intervention.


\begin{acks}
\label{sec:acknowledgement}
This material is based upon work supported by the National Science Foundation under Grant No. CNS-1837652. This material is based upon work supported by the U.S. Department of Energy’s Office of Energy Efficiency and Renewable Energy (EERE) under the Vehicle Technologies Office award number CID DE-EE0008872.  The views expressed herein do not necessarily represent the views of the U.S. Department of Energy or the United States Government. 
\end{acks}

\bibliographystyle{ACM-Reference-Format}
\bibliography{ICCPS2021}

\newpage
\section{Appendix}
\balance

This section contains an analysis of a stock ACC system tested under the same protocol as the Dynamic Time-gap, and comments on cut-ins that occur during data collection.
\subsection{ACC Dynamic Time-gap Test}

    \begin{figure}
    \centering
    \includegraphics[width=\linewidth]{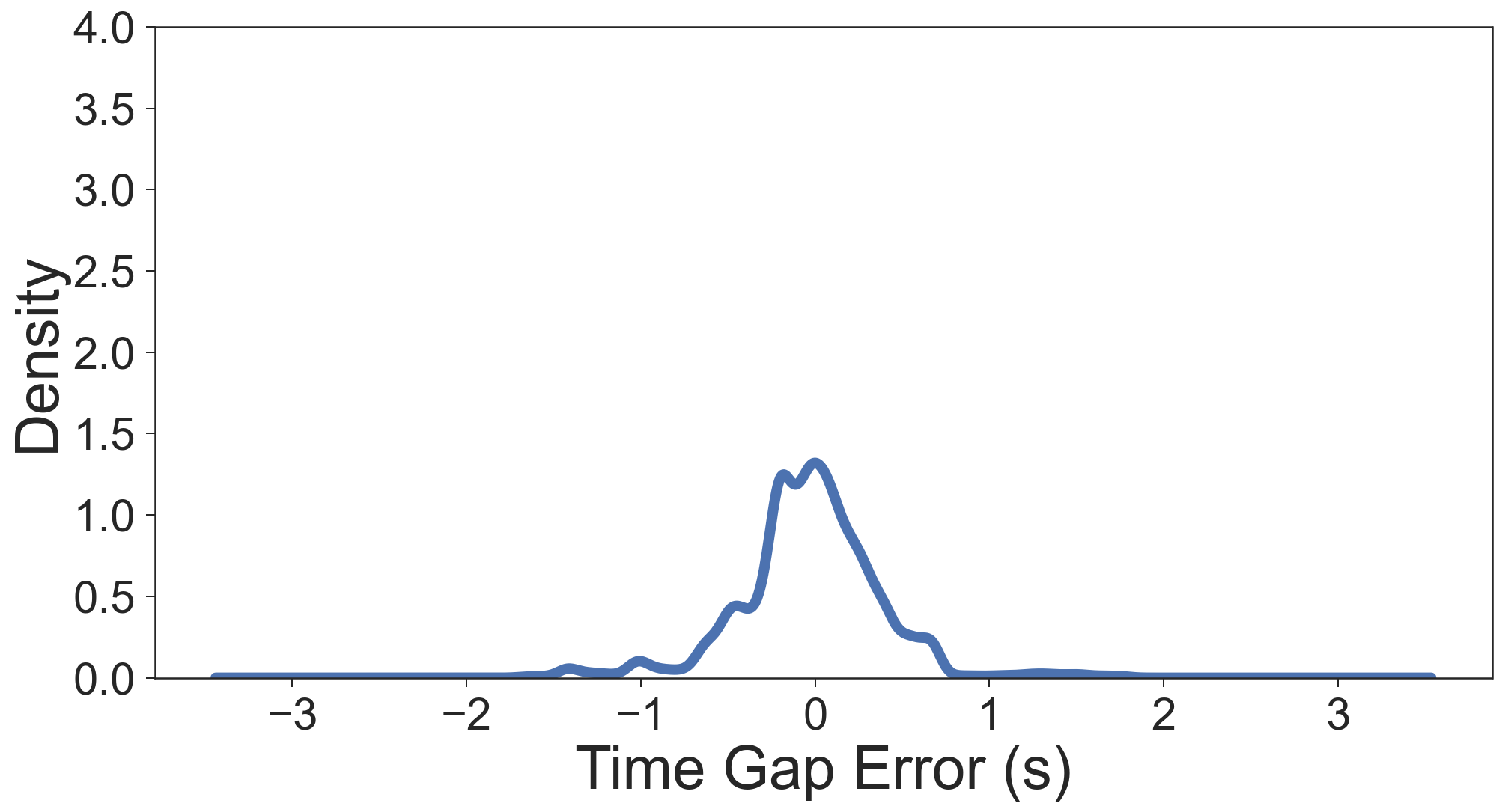}
        \caption{Time-gap error distributions for the stock ACC system under Dynamic Time-gap control. This distribution includes cut-in events.}
        \label{fig:acc}
    \end{figure} 

Though the aim of this work is not to recreate ACC with human drivers, it is interesting to compare between ACC driving and human driving with CAN Coach. In Figure~\ref{fig:acc} we can see how the stock ACC system compares to the human drivers with CAN Coach in Figure~\ref{fig:dthccerror}.

The same dynamic time-gap protocol was run, but this time with the stock ACC active. The ACC has a mean time gap error of -0.03 s and a standard deviation of 0.40 s.
As a side note, the ACC error calculation is based on an assumed constant time-gap strategy of the stock ACC system and the stated time-gap settings in the owner's manual of the vehicle.
The ACC driving shows a smaller mean error (0.03 s < 0.08 s), and a higher standard deviation (0.40 s > 0.30 s) in comparison to the average human driver under Dynamic Time-gap Coached Driving. 


\subsection{Discussion on Cut-Ins}

The planned experiments were not structured to account for the semi-random nature of vehicle cut-ins. If, for example, during one driver's test there were two cut-in events and during another driver's test there were none, the results could be skewed. Broadly, the number and nature of cut-ins across drivers could be very different and bias the results. It would be interesting to do an analysis testing the velocity and spacing thresholds for cut-ins, and even make an online predictive model based on the state-space sensed by a vehicle in real-time.

In practice, the number of cut-ins were minimal. Each driver had 2.6 cut-in events on average over the course of the nearly hour-long drives. Some of these cut-ins are during the Ghost Mode, where the driver is following a Ghost Car. In this setting the cut-in is even more abstract: there is a real vehicle cutting in front of the CAN Coach equipped vehicle which is following a virtual vehicle. Drivers were not given any specific guidance on how to respond to cut in events, and the sample size is small by construction (i.e., the selection of a remote freeway). Consequently we do not analyze the cut in data rigorously in this article, rather we discard the data surrounding the events.

\end{document}